\pdfoutput=1
\documentclass[11pt]{article}
\usepackage[final]{acl}                  %
\usepackage{times}
\usepackage{latexsym}
\usepackage[T1]{fontenc}
\usepackage[utf8]{inputenc}
\usepackage{microtype}
\usepackage{inconsolata}
\usepackage{graphicx}
\usepackage{booktabs}
\usepackage{tabularx}
\usepackage{amsmath}
\usepackage{xcolor}
\usepackage{float}
\usepackage{enumitem}
\usepackage{stfloats}                    %

\newcommand{\lw}{LATTE-WordSeg-Lite}      %
\newcommand{\ulong}{\texttt{long}}        %
\newcommand{\ushort}{\texttt{short}}      %
\newcommand{\umin}{\texttt{minimal}}

\newcommand{\AvgWord}{85.65}
\newcommand{\AvgSefr}{84.52}
\newcommand{\AvgDelta}{$+1.13$ [$+0.80$, $+1.52$]}
\newcommand{\AvgComWord}{87.80}
\newcommand{\AvgComOther}{83.78}
\newcommand{\AvgComDelta}{$+4.01$ [$+3.43$, $+4.69$]}
\newcommand{\AvgUnseenWord}{81.88}
\newcommand{\AvgUnseenOther}{81.16}
\newcommand{\AvgUnseenDelta}{$+0.72$ [$+0.36$, $+1.10$]}
\newcommand{\PudShortDeep}{$-0.96$ [$-1.54$, $-0.36$]}
\newcommand{\OrchidDictMore}{0 to 2}
\newcommand{\OrchidNeuralMore}{6 to 26}
\newcommand{\NlpoNewmmMax}{1.9}

\newcommand{\WsOskut}{91.3}
\newcommand{\WsSefr}{86.7}
\newcommand{\OskutWsElsewhere}{64.4--81.0}
\newcommand{\SefrWsElsewhere}{74.8--85.2}

\newcommand{\FourOnBest}{95.8--97.6}
\newcommand{\FourElsewhere}{75.9--85.6}

\newcommand{\KhammoonKhm}{98.3}
\newcommand{\KhammoonElsewhere}{68.0--85.5}
\newcommand{\UnitSpreadKhm}{26.9}

\newcommand{\OskutWsVistec}{71.5}

\newcommand{\DictAvg}{70.9}
\newcommand{\DictGap}{13}

\newcommand{\ExposedKhm}{9.1\%}

\newcommand{\ExposedKhmStudent}{8.0\%}

\newcommand{\ExposedMaxOtherStudent}{3.1\%}
\newcommand{\ExposedChangeAll}{1.27}
\newcommand{\ExposedChange}{0.49}
\newcommand{\DeepAheadRange}{3.1--15.6}
\newcommand{\DeepBestGap}{10.6}
\newcommand{\KhammoonKhmDelta}{$+0.12$ [$+0.08$, $+0.16$]}

\newcommand{\AvgNewmm}{71}
\newcommand{\AvgDeepcutRound}{84}
\newcommand{\SpeedLite}{97--105K}
\newcommand{\SpeedKhammoon}{40--43K}
\newcommand{\SpeedDeepcut}{2.6K}

\newcommand{\SpeedNewmm}{0.9M}
\newcommand{\SpeedAttacut}{78--100K}
\newcommand{\SpeedTeacherCpu}{66}
\newcommand{\SpeedVsKhammoon}{2.4}
\newcommand{\SpeedVsSefr}{55}
\newcommand{\SpeedLineVsOskut}{55}
\newcommand{\SpeedNewmmVs}{16}
\newcommand{\SpeedNlpoVs}{85}
\newcommand{\SpeedAttaVsLine}{1.3 and 1.7}

\newcommand{\SpeedVsTeacher}{1{,}600}
\newcommand{\MemLite}{136}

\newcommand{\SpeedLiteLine}{60K}
\newcommand{\SpeedLineVsDeepcut}{23}
\newcommand{\SpeedLiteLineR}{62K}
\newcommand{\SpeedLiteBatchR}{110K}
\newcommand{\SpeedDeepcutR}{2.8K}
\newcommand{\SpeedLineVsDeepcutR}{22.5}
\newcommand{\SpeedKhammoonLine}{39K}

\newcommand{\TfCores}{1.27--1.36}
\newcommand{\TeachCrfGap}{0.1 to 0.9}
\newcommand{\ScorerScores}{98}
\newcommand{\ScorerAgree}{0.001}
\newcommand{\ScorerAgreeTltk}{0.10}
\newcommand{\NcBest}{97.94}
\newcommand{\NcOrchid}{91.52}
\newcommand{\NcPud}{87.71}
\newcommand{\SpeedLiteFp}{61K}
\newcommand{\BatchLinesMax}{2.9}
\newcommand{\BatchFoneMax}{0.05}

\newcommand{\SpeedLiteLater}{111K}
\newcommand{\SpeedKhammoonLater}{46K}
\newcommand{\BAvgWord}{93.30}
\newcommand{\BAvgSefr}{92.60}
\newcommand{\BAvgDelta}{$+0.70$ [$+0.52$, $+0.90$]}

\newcommand{\CharAllLo}{0.15}
\newcommand{\CharAllHi}{5.2}

\newcommand{\VistecShare}{72\%}
\newcommand{\VistecSeen}{92.35}
\newcommand{\VistecUnseen}{90.37}
\newcommand{\VistecGap}{$+1.98$ [$+1.66$, $+2.35$]}
\newcommand{\VistecOtherGap}{0.25}
\newcommand{\VistecCleanLines}{1{,}821}
\newcommand{\VistecNeither}{984}
\newcommand{\VistecNoHashLines}{874}
\newcommand{\VistecNoHashDelta}{$+0.95$ [$+0.72$, $+1.19$]}
\newcommand{\BestHeldDeepcut}{0.93}
\newcommand{\BestOursDeepcut}{0.96}

\newcommand{\BestHeldNewmm}{0.67}
\newcommand{\BestOursNewmm}{0.70}
\newcommand{\LooMinLead}{0.05}

\newcommand{\VistecCleanDelta}{$+2.82$ [$+2.59$, $+3.09$]}

\newcommand{\TeachVistecGap}{0.9}
\newcommand{\FidFone}{98.7}
\newcommand{\FidLines}{58\%}
\newcommand{\FidTeacher}{53\%}
\newcommand{\FidStudent}{47\%}
\newcommand{\FidVistec}{64\%}
\newcommand{\FidStudentRange}{57--65\%}

\newcommand{\SeedMaxDiff}{0.3}
\newcommand{\SeedMaxExact}{0.29}
\newcommand{\SeedMeanMax}{0.05}
\newcommand{\UoneLoss}{0.04--0.11}
\newcommand{\UoneSetMax}{0.5}
\newcommand{\GoldVistec}{0.85}
\newcommand{\GoldTud}{1.23}
\newcommand{\GoldKhm}{0.15}
\newcommand{\GoldWs}{2.17}
\newcommand{\GoldWsCi}{$-2.15$ [$-4.15$, $+0.02$]}
\newcommand{\GoldLossLarge}{2.1}

\newcommand{\GoldLossSub}{0.4}
\newcommand{\NolexGainAvg}{$+0.19$ [$+0.02$, $+0.36$]}
\newcommand{\NolexGainAvgRound}{0.2}
\newcommand{\NolexGainVistec}{1.3}
\newcommand{\NolexGainOrchid}{0.5}
\newcommand{\NolexGainBest}{0.2}
\newcommand{\NolexCostPud}{0.35}
\newcommand{\NolexUnitLong}{0.3}
\newcommand{\NolexUnitShort}{0.2}
\newcommand{\NbestOne}{85.64}
\newcommand{\NbestThree}{88.74}
\newcommand{\NbestFive}{89.92}
\newcommand{\NbestTen}{91.26}
\newcommand{\NbestTudOne}{39}
\newcommand{\NbestTudThree}{60}
\usepackage{url}
\makeatletter                            %
\renewcommand\paragraph{\@startsection{paragraph}{4}{\z@}%
  {0.75ex plus 0.3ex minus 0.1ex}{-1em}{\normalsize\bfseries}}
\makeatother
\title{LATTE-WordSeg-Lite: Multi-Granularity Thai Word\\Segmentation on a CPU, with a Reproducible Benchmark}

\author{Thodsaporn Chay-intr$^{1,2,3}$ \enspace Kobkrit Viriyayudhakorn$^{1,2,4}$ \enspace Thanaruk Theeramunkong$^{3,5}$ \\
  $^{1}$iApp Technology, Thailand \quad $^{2}$OpenThai Laboratory, Thailand \\
  $^{3}$Intelligent Informatics and Service Innovation Research Center, Thailand \\
  $^{4}$Artificial Intelligence Entrepreneur Association of Thailand (AIEAT), Thailand \\
  $^{5}$Sirindhorn International Institute of Technology, Thammasat University, Thailand \\
  \texttt{t.chayintr@gmail.com, kobkrit@aieat.or.th, thanaruk@siit.tu.ac.th}}

\begin{document}
\maketitle

\begin{abstract}
Large language models read Thai through subword tokens, without a word segmenter. Yet Thai treebanks, speech
metrics and morphological resources are built on words, often at several granularities, and word boundaries
resolve ambiguities that subword tokens leave open. LATTE (Lattice ATTentive Encoding) reads a text at several
granularities at once, through a lattice of dictionary words, but it is a large GPU model. We distil it into
\lw{}, a CPU segmenter of 1.17M parameters that keeps the lattice as features. It outputs compounds, words and
morphemes in one pass, and learns only from annotated data and text whose licences allow commercial use. To
compare it with other tools, published Thai scores do not suffice, as they come from different files, splits and
scorers. We therefore rebuild seven test sets from pinned sources and score twelve systems under one protocol.
With the unit that fits each test set, \lw{} scores above DeepCut on five sets, four of them from corpora that
its teachers learnt from. Called line by line on one CPU thread, it runs \SpeedLineVsDeepcut{} times as fast as
DeepCut. We release the model, the benchmark scripts and all predictions.
\end{abstract}

\section{Introduction}

\begin{table*}[t]\centering\small
\begin{tabular*}{\linewidth}{@{\extracolsep{\fill}}lrrrrrrrrr@{}}
\toprule
 & \multicolumn{4}{c}{Commercial-use licence} & \multicolumn{3}{c}{Non-commercial licence} & \multicolumn{2}{c}{Average} \\
\cmidrule(lr){2-5}\cmidrule(lr){6-8}\cmidrule(l){9-10}
System & TUD & PUD & WS160 & VISTEC & BEST & ORCHID & khm & All & Commercial \\
\midrule
\textbf{LATTE-WordSeg-Lite} & & & & & & & & & \\
\quad \texttt{long} & 83.71 & \textbf{86.18} & 85.51 & \textbf{92.77} & 82.09 & 79.77 & 71.54 & 83.08 & \underline{87.04} \\
\quad \texttt{short} & \textbf{90.27} & 84.66 & \underline{88.32} & 87.94 & 86.97 & 79.10 & 82.29 & \textbf{85.65} & \textbf{87.80} \\
\quad \texttt{minimal} & 85.36 & 80.34 & 84.08 & 76.48 & 82.09 & 67.86 & \textbf{98.44} & 82.09 & 81.57 \\
\midrule
khammoon & \underline{85.52} & 79.90 & 82.87 & 75.68 & (82.17) & 67.97 & \underline{98.33} & 81.78 & 80.99 \\
DeepCut & 85.50 & \underline{85.62} & 83.13 & 80.12 & \underline{(97.54)} & 76.68 & 82.81 & 84.49 & 83.59 \\
SEFR CUT-best & 85.42 & 85.55 & 83.11 & 80.39 & \textbf{(97.55)} & 76.78 & 82.82 & \underline{84.52} & 83.62 \\
SEFR CUT-ws1000 & 85.21 & 84.77 & (86.69) & 78.48 & (94.22) & 74.84 & 83.95 & 84.02 & 83.78 \\
AttaCut-SC & 85.32 & 85.34 & 83.04 & 78.75 & (96.43) & 76.44 & 82.87 & 84.03 & 83.12 \\
AttaCut-C & 85.03 & 84.44 & 81.89 & 77.29 & (95.79) & 75.91 & 82.35 & 83.24 & 82.16 \\
OSKut-ws & 74.97 & 68.70 & \textbf{(91.28)} & 71.46 & (75.19) & 64.43 & 80.99 & 75.29 & 76.60 \\
OSKut-scads & 78.01 & 83.09 & 83.20 & \underline{(89.69)} & (78.68) & 73.64 & 69.75 & 79.44 & 83.50 \\
TLTK & 76.51 & 73.09 & 77.63 & 79.92 & 75.29 & \textbf{80.68} & 65.40 & 75.50 & 76.79 \\
PyThaiNLP nlpO3 & 69.70 & 67.90 & 73.01 & 75.00 & 69.46 & \underline{80.63} & 60.82 & 70.93 & 71.40 \\
PyThaiNLP newmm & 69.64 & 67.96 & 73.35 & 76.88 & 69.13 & 79.17 & 60.18 & 70.90 & 71.96 \\
\bottomrule
\end{tabular*}

\caption{Word F1 (\%) of twelve systems under one protocol (\S\ref{sec:bench}). The test sets are grouped by
licence. The averages weigh each set equally, over all seven sets and over the four whose licence allows commercial use. The three rows of
\lw{} are the three units of one network. Its teachers learnt from the corpora of TUD, WS160, VISTEC and khm,
without their test lines. Bold marks the highest score of a column, and underline the second highest. A score is
in parentheses when the system has seen most of the test lines of that set in training (72 to 100\%). Part of
such a score may come from memorised lines (\S\ref{sec:bench}), and Section~\ref{sec:shows} measures how much.}
\label{tab:main}
\end{table*}

Large language models read Thai through subword tokens, without a word segmenter. Word segmentation can thus look
like a task of the past, as it did for neural models of Chinese~\citep{li2019necessary}. Yet word boundaries carry
information that subword tokens do not.
Japanese pre-trained models do better with a morphological analyser before the subword
tokeniser~\citep{fujii2023tokenizers}. In Chinese, up to a quarter of word boundaries fall inside a token, and
this mismatch misleads current models~\citep{wang2024tokenization,xu2026partial}. Asking a large model to segment
is no shortcut, as a prompted GPT-4, which knows no convention, reaches 75--76 token F1 on Thai segmentation~\citep{zhang2025segment}. Words also remain the units of treebanks, of the
word error rate of speech recognisers and of morphological
resources~\citep{demarneffe2021ud,phatthiyaphaibun2022wav2vec,khammoon2026}. These resources use words at
several granularities, from compounds to words and morphemes.

\begin{figure}[t]\centering
\includegraphics[width=\columnwidth]{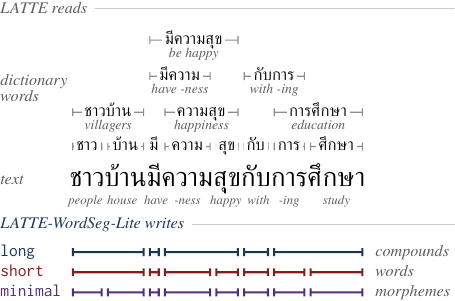}
\caption{The idea, on the phrase ``villagers are happy with education''. A bar marks the characters of one word.
The glosses are literal, and \emph{-ness} and \emph{-ing} stand for two prefixes that form nouns. LATTE reads the
characters with every dictionary word that matches them. The phrase has 33 such words, and 14 are drawn. Some
are not words of this phrase. \lw{} keeps this input and writes three units from one small network.}
\label{fig:idea}
\end{figure}

LATTE, Lattice ATTentive Encoding~\citep{chayintr2023latte}, brings several granularities to the input. It was
proposed for
languages written without spaces between words, and evaluated on Thai, Chinese and Japanese. A string in such a
language often has several readings, so LATTE reads a text at several granularities at once
(Figure~\ref{fig:idea}). It joins the characters and every dictionary word that matches them in a lattice, and a
graph attention network encodes this lattice over a pre-trained encoder.

\begin{figure}[t]\centering
\includegraphics[width=\columnwidth]{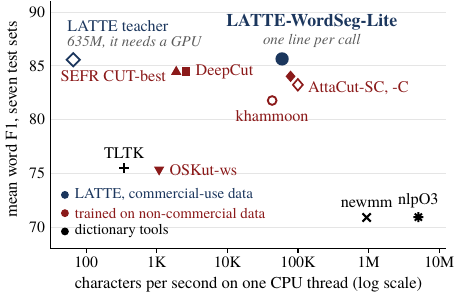}
\caption{Accuracy against speed on one CPU thread (\S\ref{sec:results}), with \lw{} at its default unit,
\ushort. khammoon segments morphemes, and six of the seven sets count words. The LATTE teacher, a GPU model, is
placed at its speed on the same CPU thread.}
\label{fig:tradeoff}
\end{figure}

In practice, however, Thai segmenters trade accuracy against speed
(Figure~\ref{fig:tradeoff}). LATTE is a GPU model, and on one CPU thread a LATTE model of this paper segments only
\SpeedTeacherCpu{} characters per second. DeepCut~\citep{kittinaradorn2019deepcut} is accurate, averaging
\AvgDeepcutRound{} word F1 over our seven test sets, but segments only \SpeedDeepcut{} characters per second.
AttaCut~\citep{chormai2019attacut} is nearly as accurate, at \SpeedAttacut{} characters per second. The
default tokeniser of PyThaiNLP, the dictionary-based \texttt{newmm}~\citep{phatthiyaphaibun2023pythainlp}, is
faster still, at \SpeedNewmm{} characters per second, but averages \AvgNewmm. khammoon~\citep{khammoon2026} is fast and
accurate on a CPU, but for morphemes.

Licences also matter. DeepCut, AttaCut, SEFR CUT and OSKut are released under the MIT licence. Yet DeepCut and
AttaCut were trained on BEST~\citep{kosawat2009best}, whose licence is non-commercial, and SEFR CUT and OSKut
build on DeepCut. The weights of khammoon are released under a non-commercial licence (CC~BY-NC-SA~4.0). We want a
segmenter that anyone may use commercially, as accurate as the best of them and fast on a CPU.

This paper brings the idea of LATTE to a CPU. \textbf{\lw{}}\footnote{\raggedright The model is on Hugging Face at \url{https://huggingface.co/iapp/OpenThai2.0-LATTE-WordSeg-Lite}, and the benchmark at \url{https://huggingface.co/datasets/iapp/OpenThai2.0-WordSeg-Benchmark}.} keeps the lattice
as its input, in the form of a few features of each character. It learns the decisions of the graph network by
distillation~\citep{hinton2015distilling} from two LATTE teachers, which we rewrote and then trained (\S\ref{sec:model}).

\lw{} also extends the idea from the input to the output. Thai corpora disagree on what a word is, so one
small network, of 1.17M parameters, writes three units in a single pass, all learnt on the same text. We name them after
the long and short unit words of the Japanese corpus BCCWJ and the smallest units they are built
from~\citep{maekawa2014bccwj}. \ulong{} keeps compounds whole, \ushort{} gives their words and \umin{} their
morphemes.

The model is trained only on annotated data and text whose licences allow commercial use, and it runs in ONNX
Runtime~\citep{onnxruntime} on one CPU thread. We release it in OpenThai~2.0, a family of models for Thai, under
the name \mbox{OpenThai2.0-LATTE-WordSeg-Lite}.

Table~\ref{tab:main} compares twelve systems (Appendix~\ref{app:systems}). With the unit that fits each test
set, \lw{} scores above DeepCut on five of the seven sets. Called line by line, it also runs \SpeedLineVsDeepcut{} times as fast on one
CPU thread. On average, it is as accurate as the labels of its teachers (\S\ref{sec:results}).

Such a comparison is only as good as its benchmark. Published Thai scores come from different files, splits and
scorers, and some training sets contain test lines. We therefore rebuild seven test sets from pinned sources. We
score twelve systems with one scorer, by the two metrics of the LATTE and khammoon papers. We also check our runs
of the other tools against their published scores (\S\ref{sec:bench}). Anyone can recompute the scores of all
twelve systems from the released predictions.

\section{Related Work}

\paragraph{Thai segmenters.} Thai segmenters are of two kinds. The first matches the text against a
dictionary. \texttt{newmm}, the default tokeniser of PyThaiNLP, uses maximum matching, and nlpO3 is an
adaptation of it in Rust~\citep{nlpo3}. TLTK joins syllables into words with a dictionary and their
collocation~\citep{tltk}. The second learns from a segmented corpus. DeepCut is a convolutional network over
characters~\citep{kittinaradorn2019deepcut}. AttaCut is a faster one, and its SC model also reads
syllables~\citep{chormai2019attacut,chormai2020syllable}. SEFR CUT and OSKut add a second model, which re-scores the
output of DeepCut and so adapts it to a new domain \citep{limkonchotiwat2020sefr,limkonchotiwat2021oskut}.
khammoon is a small network for a CPU that segments morphemes, and our student follows its layout. Each model of
these tools follows one convention, that of the corpus or the dictionary it was built from.

\paragraph{Conventions and granularities.} Corpora disagree on what a word is, and two lines of work accept
this. Multi-criteria models learn the conventions of several corpora in one network that is told which
convention to write~\citep{qiu2020concise,huang2020fast}. Multi-grained segmentation gives one sentence several granularities
at once, with small words inside larger ones. \citet{gong2017multigrained} built such data from three corpora
of different conventions. Japanese resources offer several granularities too. BCCWJ
is annotated in short and long unit words, and the tokeniser Sudachi splits in three modes
\citep{maekawa2014bccwj,omura2021word,takaoka2018sudachi}. \lw{} follows both lines for Thai, as each corpus
convention becomes one of its granularities.

\paragraph{Lattices.} A character model reads a sentence character by character and does not see its words. A
lattice model adds them, as every dictionary word that matches the sentence becomes a node over the characters
that it spans~\citep{zhang2018lattice,li2020flat}. For Thai, an earlier model lets each character attend to its candidate
words and character clusters~\citep{chayintr2021character}. LATTE followed it, and encodes the lattice of a
sentence with a graph attention network.

The words of a lattice can also reach a model as features of each character, which is
faster~\citep{ma2020simplify}. Before these models, a pointwise Japanese segmenter already marked, by length, whether
a dictionary word lies left of, right of or over a boundary~\citep{neubig2011pointwise}. So do our features.

\paragraph{Small students.} A large model can teach a small one. In distillation, a small student learns from
the outputs of a large teacher, and not from gold labels alone~\citep{hinton2015distilling}. This has made a
Chinese segmenter faster~\citep{huang2020fast} and a Japanese analyser smaller~\citep{tolmachev2019shrinking},
and \lw{} does both for Thai.

\paragraph{Evaluation.} A score depends on more than the model. Rankings on a standard split may not hold on
other splits~\citep{gorman2019standard}. Random splits can overestimate performance~\citep{sogaard2021random}.
Test items that repeat training items reward memorisation, not
generalisation~\citep{lewis2021overlap,elangovan2021memorization}. For Thai, \citet{chormai2019attacut} compared
the tokenisers of their time on three domains with one scorer. \citet{fu2020rethinkcws} revisited Chinese
segmentation in this spirit, and we revisit Thai word segmentation.

\begin{figure*}[t]\centering
\includegraphics[width=\textwidth]{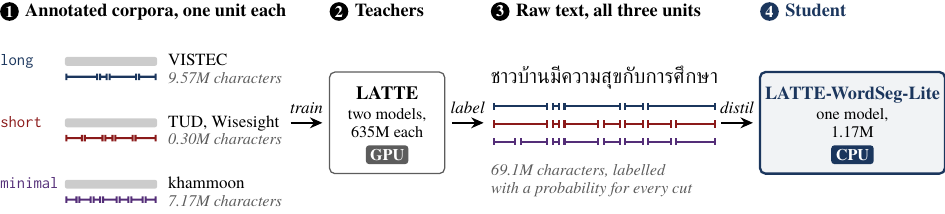}
\caption{How \lw{} is made. A grey bar is a line of text, and the coloured bars under it are its words in one unit.
Each corpus is annotated in one unit. The teachers label raw text with all three units, here
the phrase of Figure~\ref{fig:idea}, and the student learns from these labels. Sections~\ref{sec:corpora} to \ref{sec:student} take the four
steps in turn.}
\label{fig:method}
\end{figure*}

\section{LATTE-WordSeg-Lite}\label{sec:model}

Figure~\ref{fig:method} shows how \lw{} is made, in four steps. The four parts of this section take one step
each, under its title in the figure.

\begin{figure}[t]\centering
\includegraphics[width=\columnwidth]{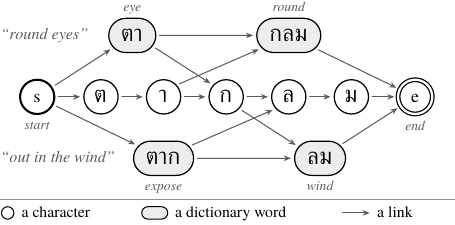}
\caption{The lattice of a string of five characters, drawn as in the LATTE paper~\citep{chayintr2023latte}. A
link joins a node to each node that begins where it ends, so each path from \emph{s} to \emph{e} is one
reading. Four of its seven dictionary words are drawn.}
\label{fig:graph}
\end{figure}

\subsection{Annotated Corpora, One Unit Each}\label{sec:corpora}

In step~1, each Thai corpus is annotated in one unit. VISTEC keeps compounds whole, our
\ulong{} unit. TUD and Wisesight-1000 split them into words, \ushort. khammoon splits words into morphemes,
\umin. Each unit is thus the convention of its corpora, and its name is approximate.

These corpora are also different texts, of 9.57M, 0.30M and 7.17M characters. So no line carries all three
units, and \ushort{} has far less data.

\subsection{Teachers}

\paragraph{Why teachers.} Step~2 trains teachers, which remove these two limits. A teacher that has learnt a
corpus can write its unit on any text. So every line can carry all three units, and each unit gets as much text as the
others. Without teachers, a network trained on the
corpora of one unit is \GoldLossSub{} to \GoldLossLarge{} points lower over the seven test sets
(\S\ref{sec:results}).

\paragraph{The lattice as a graph.} Our teachers are LATTE models, and LATTE sees a text as a graph
(Figure~\ref{fig:graph}). Each character is a node, and so is each dictionary word that matches the text. A link
joins two nodes when one ends where the other begins. A path of links from the start of the text to its end is
thus one reading. Graph attention passes information along the links in both directions. Each character then
gathers the words that cover it, and a CRF segments the text.

\paragraph{Rewriting LATTE.} We rewrote LATTE for three reasons (Appendix~\ref{app:training}). The first is
\emph{speed}. Our code takes a lattice word from the encoder states of its characters and batches the lattice
attention, so a training step is 3.6 times as fast at the same precision. The second is
\emph{conventions}. A corpus embedding and one CRF per corpus let one model learn each corpus in its own
convention, which gives \lw{} its units. The third is \emph{soundness}. The dictionary holds the training
vocabulary. So a word known only from training data is hidden from the lattice of its own sentence, as at test
time.

\paragraph{Two teachers.} We trained two such models on XLM-RoBERTa-large \citep{conneau2020xlmr}. Each has 635M
parameters and trains in under an hour on one H100 GPU. Teacher~1 learns \ulong{} and \ushort{} from VISTEC,
TUD and Wisesight-1000 without the WS160 lines. Teacher~2 learns \umin{} from 4{,}763 khammoon documents
licensed CC~BY-SA.

\subsection{Raw Text, All Three Units}\label{sec:rawtext}

\paragraph{Text.} In step~3, the teachers label 69.1M characters of raw text in all three units. The text comes
from Thai Wikipedia, the Wisesight sentiment corpus and the training documents of Teacher~2. Before labelling, we drop the lines that contain a
test sentence of TUD, PUD, WS160, VISTEC or BEST. The filter missed ORCHID, which it read in its tagged form, and
the test set of khammoon. Appendix~\ref{app:sources} counts the test lines left.

\paragraph{Soft labels.} A label is soft. It is the probability that a
word starts at a character. \ulong{} comes from the VISTEC output
of Teacher~1, and \umin{} from Teacher~2. \ushort{} mixes the TUD and Wisesight outputs of Teacher~1, 0.6 to 0.4.
For the phrase of Figure~\ref{fig:idea}, the teachers are sure everywhere but once, where a \ushort{} boundary
inside ``villagers'' gets a probability of 0.23.

\begin{figure*}[t]\centering
\includegraphics[width=\textwidth]{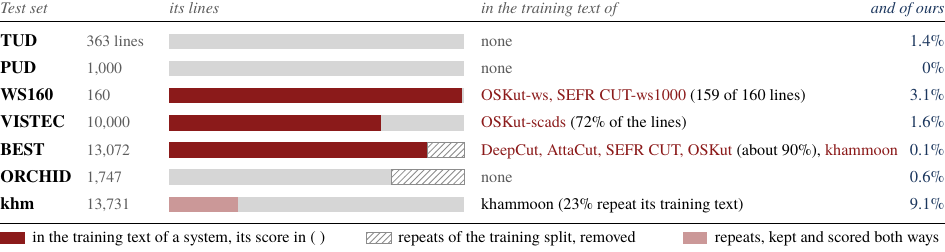}
\caption{What happened to the lines of each test set, with the number of lines kept. The scores of the systems named in dark red are in
parentheses in Table~\ref{tab:main}. Grey lines are in the training text of no system of the table. The right
column gives the share of lines with a copy or near-copy in the training text of our models
(Appendix~\ref{app:sources}).}
\label{fig:bench}
\end{figure*}

\subsection{Student}\label{sec:student}

\paragraph{Network.} In step~4, the student learns from these probabilities alone, and
no label of the corpora themselves is among its targets. It follows the layout of khammoon, with character embeddings, convolutions of width 3, 5 and
7, and a two-layer BiLSTM, 1.17M parameters in all. It reads the lattice of LATTE as 13 yes-or-no features per
character (the bars of Figure~\ref{fig:idea}). Six say whether a dictionary word starts at the character, one per
band of word length, and six whether one ends there. The last says whether the character lies strictly inside a
dictionary word (Appendix~\ref{app:training}). The 13 bits thus keep the ends and the lengths of lattice words.

\paragraph{Training.} The student has one output per unit and learns the probabilities of the three units at
once, by binary cross-entropy. It trains for 60K steps of up to 32{,}768 characters, 26 passes over the text. It ships
as one int8 ONNX graph of 3~MB, with a dictionary of 0.8~MB beside it.

\paragraph{Decoding.} A word starts where the logit is non-negative and Thai spelling allows it
(Appendix~\ref{app:training}). Every \ulong{} word is made of whole \ushort{} words, and never cuts one. \umin{} is decoded on its own, as its convention keeps together some punctuation that \ushort{}
splits. It still shares 99.1\% of the \ushort{} boundaries on our test sets.

\paragraph{Licences.} The licences of all annotated data, raw text and word lists allow commercial use, as their
sources state them (Appendix~\ref{app:training}). No text or labels were taken from BEST or ORCHID for the
teachers, the dictionary or the student. Two indirect links remain. khammoon corrected its LLM labels with a
model fine-tuned on its own human labels of BEST sentences. Still, it releases the labels of each row under the
licence of that row, and we use only its rows under CC~BY-SA. The encoder of the teachers, XLM-RoBERTa, was
itself pre-trained on web text. Our weights and dictionary are CC~BY-SA~4.0, and our code is Apache-2.0.

\section{Revisiting the Benchmarks}\label{sec:bench}

A score can be compared only if three things are known. They are its test file, the training data of the system
and its scorer. Published Thai results often leave one open, so we take them in turn.

\begin{table}[t]\centering\small\setlength{\tabcolsep}{4pt}
\begin{tabular*}{\linewidth}{@{\extracolsep{\fill}}llrrl@{}}
\toprule
Set & Text & Lines & Words & Unit \\
\midrule
TUD & Wikipedia, TNC & 363 & 7.7k & \texttt{short} \\
PUD & news, Wikipedia & 1,000 & 22.3k & \texttt{long} \\
WS160 & social media & 160 & 3.4k & \texttt{short} \\
VISTEC & social media & 10,000 & 677k & \texttt{long} \\
BEST$^{\dagger}$ & four genres & 13,072 & 487k & \texttt{short} \\
ORCHID$^{\dagger}$ & technical papers & 1,747 & 24.1k & \texttt{long} \\
khm$^{\dagger}$ & instructions & 13,731 & 426k & \texttt{minimal} \\
\bottomrule
\end{tabular*}

\caption{Test sets. TNC is the Thai National Corpus. BEST and ORCHID are de-duplicated. Unit is the unit of
\lw{} closest to the words of a set (\S\ref{sec:bench}). $^{\dagger}$The non-commercial sets are
rebuilt by script and are not redistributed (Appendix~\ref{app:sources}).}
\label{tab:sets}
\end{table}

\subsection{Test Sets}

\paragraph{Sources.} The first condition is the test file. We use seven test sets (Table~\ref{tab:sets}). TUD and PUD are two Thai treebanks of
Universal Dependencies \citep{demarneffe2021ud,sriwirote2025tud,zeman2017conll}. WS160
\citep{suriyawongkul2024wisesight} and VISTEC, short for VISTEC-TP-TH-2021 \citep{limkonchotiwat2021oskut}, are
social media. khm is the human test set of khammoon, 2{,}000 instruction documents
\citep{limkonchotiwat2025wangchan}. The official test set of BEST has no gold boundaries in its public release,
and ORCHID~\citep{sornlertlamvanich1999orchid} has no test split. We therefore split both 80/10/10 ourselves.

\paragraph{Cleaning.} We then clean the sets in three ways. ORCHID stores punctuation as tags, so we rebuild its
sentences as natural text. In every set, we let a space always separate words. Last, lines repeat between the
training and test text of a corpus.
The web address of a newspaper fills 136 test lines of BEST, and the heading ``Abstract'' 15 of ORCHID. We remove
such repeats from these two sets, whose splits are ours. For khm,
we report both scores, with and without its repeats of the training text of khammoon (the set khmdd).

\paragraph{Rebuilding.} Scripts repeat these steps. They download each source at a pinned revision, rebuild the
test sets and check every file against its SHA-256 (Appendix~\ref{app:sources}). We release the four open test files, and all predictions as word offsets.

\subsection{Lines Seen in Training}

The second condition is the training data. Figure~\ref{fig:bench} shows, for each test set, the lines that some
system has seen in training. Wisesight-1000, on which two engines were trained, contains 159 of the 160 WS160
lines. The split on which the VISTEC engine of OSKut was trained contains \VistecShare{} of the VISTEC test lines
of the corpus repository. Tools trained on BEST, and khammoon, have seen most or all of our BEST test lines, as
our split is random (Appendix~\ref{app:systems}). We keep such scores, and put them in parentheses when the
training text holds most of the test lines (72 to 100\%). Part of such a score may come from memorised lines,
which \S\ref{sec:shows} measures.

\begin{figure}[t]\centering
\includegraphics[width=\columnwidth]{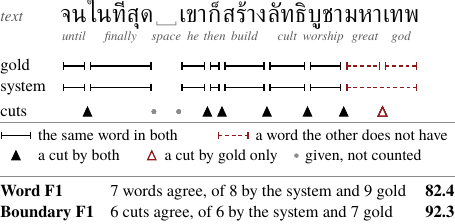}
\caption{One line of TUD with literal glosses (``until finally he founded a cult to worship the great god''), and
the output of \texttt{newmm} as the system. The two cuts next to the space are given to every system, and
boundary F1 therefore does not count them.}
\label{fig:score}
\end{figure}

\subsection{Scoring}

\paragraph{Conventions.} The third condition is the scorer, which must know what a word is. The corpora
cut the same word differently. \emph{kanthamngan}, ``working'', is usually one word in VISTEC, two in ORCHID
(\emph{kan\,$|$\,thamngan}) and three in BEST and khm (\emph{kan\,$|$\,tham\,$|$\,ngan}). Each set is
therefore scored in its own convention. A system with several units uses the unit trained on that convention,
or else the unit chosen on held-out data of the same corpus (Appendix~\ref{app:units}).
Table~\ref{tab:sets} names it for each set.

\paragraph{Two metrics.} Each set is then scored with the two metrics of the LATTE paper (Figure~\ref{fig:score}). \emph{Word F1}
is its word-level F1. A word of the system, cut at every space, is correct when a gold word has the same start
and the same end.

\emph{Boundary F1} is the character-level F1 of that paper. A cut between two characters is correct when gold
cuts there too. A space separates words, so the cuts next to a space, and the start of a line, are given
to every system. We leave them out, as the metric of khammoon does.

In Figure~\ref{fig:score}, one missing cut costs one boundary but three words, so boundary F1 is usually the
higher of the two. Both metrics sum their counts over a set. Differences carry 95\% intervals from a paired
bootstrap~\citep{koehn2004significance,bergkirkpatrick2012empirical}, with 1{,}000 resamples of the
lines.

\paragraph{Published scores.} Last, we check our runs of the other tools. Each runs with its released model. To
check that we run it as its authors did, we score our outputs their way, on their test files
(Appendix~\ref{app:published}). OSKut and DeepCut match the OSKut paper within 0.01. SEFR CUT and
four of the five tools of the khammoon table match their papers to the decimal printed.
AttaCut and \texttt{newmm} match less well.

Where a score is reproduced, Table~\ref{tab:main} differs from it because files and scorers differ,
not because the tool ran differently. For example, one output of DeepCut on WS160 has a word F1 of 84.03 when a space counts as
a word, and 80.42 as a mean over lines. Our scorer gives 83.13, as in Table~\ref{tab:main}.

\section{Results}\label{sec:results}

\subsection{Accuracy and Speed}

\paragraph{Accuracy.} Each unit has the highest score on one set of its convention (Table~\ref{tab:main}).
These are TUD for \ushort, VISTEC for \ulong{} and khm for \umin. On the other four sets, other systems are level or ahead.
On PUD, \ulong{} is level with DeepCut and SEFR CUT-best.
On WS160 and BEST, every system above \ushort{} was trained on the test lines. On ORCHID, \ulong{} is 0.9 below
nlpO3 and TLTK.

For each set, take the unit of its convention, or the unit chosen on held-out data (\S\ref{sec:bench}). Then
\lw{} is above DeepCut on five sets, by \DeepAheadRange{} points. It is level on PUD and \DeepBestGap{} below on
BEST (Appendix~\ref{app:units} gives all intervals). With \ushort{} on every set, it is above DeepCut on four sets and
below on PUD, khm and BEST. On the test of khammoon, \umin{} is within 0.2 points of khammoon (\KhammoonKhmDelta), whose
labels it learnt from. On the other six sets, khammoon is out of convention, as it gives morphemes and not the words that these sets count.

With \ushort{} on every set, \lw{} averages \AvgWord, the highest average of the table. SEFR CUT-best is second with
\AvgSefr, a difference of \AvgDelta. \lw{} keeps the highest average when any one set is left out, though by
only \LooMinLead{} without VISTEC. Over the four sets whose licence allows commercial use, it averages
\AvgComWord, against \AvgComOther{} for SEFR CUT-ws1000 (\AvgComDelta). Boundary F1 puts the same two systems at the top
(Appendix~\ref{app:units}).

These averages include TUD, WS160, VISTEC and khm, whose corpora the teachers learnt from. On PUD and
ORCHID, which no system learnt from, \ushort{} averages \AvgUnseenWord, against \AvgUnseenOther{} for SEFR
CUT-best (\AvgUnseenDelta). Some design choices also looked at test lines of four sets, so our scores on them may be optimistic (Limitations). We do not
train on BEST and ORCHID. A LATTE model trained also on them, LATTE-NC below,
reaches \NcBest{} and \NcOrchid{} on them, and \NcPud{} on PUD. It is not one of our teachers.

\begin{table}[t]\centering\small
\begin{tabular*}{\linewidth}{@{\extracolsep{\fill}}lrr@{}}
\toprule
System & Chars/s & Our model is \\
\midrule
PyThaiNLP nlpO3 & 5.06M & 85$\times$ slower \\
PyThaiNLP newmm & 937K & 16$\times$ slower \\
\textbf{LATTE-WordSeg-Lite}, batched & 105K & -- \\
AttaCut-C & 100K & 1.7$\times$ slower \\
AttaCut-SC & 78K & 1.3$\times$ slower \\
\textbf{LATTE-WordSeg-Lite}, per line & 60K & -- \\
khammoon & 43K & 2.4$\times$ faster \\
DeepCut & 2.6K & 23$\times$ faster \\
SEFR CUT-best & 1.9K & 55$\times$ faster \\
OSKut-ws & 1.1K & 55$\times$ faster \\
TLTK & 345 & 170$\times$ faster \\
\midrule
LATTE teacher, CPU & 66 & 1{,}600$\times$ faster \\
LATTE teacher, GPU & 92K & 1.1$\times$ faster \\
\bottomrule
\end{tabular*}

\caption{Characters per second on one CPU thread. All but the teacher segment
the same 768 lines, model loading left out (Appendix~\ref{app:systems}). The last column compares each tool with
our row called the same way, batched for khammoon, SEFR CUT and the teacher.}
\label{tab:speed}
\end{table}

\paragraph{Speed.} Table~\ref{tab:speed} compares each tool with \lw{} called the same way, on one CPU thread.
khammoon and SEFR CUT take all lines in one call. Called so, \lw{} segments \SpeedLite{} characters per second,
with \MemLite{}~MB of memory (Appendix~\ref{app:systems}). It is \SpeedVsKhammoon{} and \SpeedVsSefr{} times as fast
as they are.

The other tools take one line per call. Called so, \lw{} segments \SpeedLiteLine{} characters per second. It is
\SpeedLineVsDeepcut{} and \SpeedLineVsOskut{} times as fast as DeepCut and OSKut. Both AttaCut models
are faster, by \SpeedAttaVsLine{} times, and so are the dictionary tokenisers \texttt{newmm} and nlpO3, by
\SpeedNewmmVs{} and \SpeedNlpoVs{} times. Yet these two average \DictAvg{} word F1, against \AvgWord.

\subsection{From Teachers to Student}

\paragraph{From GPU to CPU.} Each teacher is a GPU model of 2.5~GB, and the student a model of 3~MB. Batched on one CPU
thread, the student is as fast as its teacher on an H100 GPU. On that thread, it is \SpeedVsTeacher{} times as
fast as its teacher (Table~\ref{tab:speed}).

\begin{table}[t]\centering\small
\begin{tabular*}{\linewidth}{@{\extracolsep{\fill}}lrrr@{}}
\toprule
Model & \texttt{long} & \texttt{short} & \texttt{minimal} \\
\midrule
LATTE teachers & 82.92 & 85.57 & 81.82 \\
\textbf{\lw{}} & 83.08 & 85.65 & 82.09 \\
\addlinespace Single-unit students & 83.04 & 85.54 & 82.02 \\
Corpus-trained networks & 80.94 & 83.50 & 81.73 \\
\addlinespace Student without lattice bits & 82.78 & 85.45 & 82.08 \\
\bottomrule
\end{tabular*}

\caption{From the teachers to the student, as mean word F1 (\%) over the seven test sets at each unit. The
teachers are scored by their labels, thresholded at 0.5, which the student learnt. The other rows are networks of
the size of the student, each from one run. A single-unit student has one output and learns the same labels.
A corpus-trained network has one output too, and learns the corpora of its unit alone
(Appendix~\ref{app:units}).}
\label{tab:student}
\end{table}

Table~\ref{tab:student} shows what this speed costs in accuracy. On average over the seven sets, it costs
nothing. At every unit, the mean of the student is no lower than that of the teacher labels. On single sets, the labels lead
by \TeachVistecGap{} on VISTEC. Decoded by their CRFs, the teachers are also \TeachCrfGap{} points above the
student on TUD, WS160 and VISTEC.

\paragraph{A student, not a copy.} Equal scores do not mean equal outputs. With each set at its unit, the
boundary F1 between the cuts of the student and those of its teachers is \FidFone. Only \FidLines{} of the test
lines are segmented identically by both. Where the two differ, gold sides with the teachers in \FidTeacher{} of
the cuts and with the student in \FidStudent{} (Appendix~\ref{app:units}).

\paragraph{Units together.} A single-unit student is no better. Its mean is
\UoneLoss{} lower at every unit (Table~\ref{tab:student}). On single sets, it differs by up to \UoneSetMax{}
points either way. The three units thus share one network at the same mean. For scale, a
second run with another seed changes no score by more than \SeedMaxDiff{} points, and no mean by more than \SeedMeanMax{} (Appendix~\ref{app:units}).

\paragraph{The lattice bits.} We trained the student without its 13 lattice bits, on the same labels. Its
mean drops by \NolexUnitLong{} at \ulong{} and \NolexUnitShort{} at \ushort, and stays
level at \umin{} (Table~\ref{tab:student}). With each set at its unit, the bits add \NolexGainAvg, with \NolexGainVistec{} on
VISTEC, \NolexGainOrchid{} on ORCHID and \NolexGainBest{} on BEST. On PUD, the student is \NolexCostPud{} better
without them, and the other three sets are level within their intervals.

\paragraph{Without teachers.} A corpus-trained network, which learns the corpora of its unit alone, holds its own
on their test sets. It is \GoldVistec{} above \lw{} on VISTEC and \GoldTud{} above on TUD. It is \GoldKhm{} below
on khm, and \GoldWs{} below on the 160 lines of WS160, within the interval (Appendix~\ref{app:units}). Over the
seven sets, however, its mean is \GoldLossLarge{} lower at \ulong{} and \ushort, and \GoldLossSub{} lower at
\umin. Teacher labels on broad raw text carry each unit beyond its own corpus. Our runs do not separate the effect of the
labels from that of the text.

\subsection{What the Benchmark Shows}\label{sec:shows}

\paragraph{Seen lines score higher.} Two engines were trained on Wisesight-1000. OSKut-ws scores \WsOskut{} on WS160
and \OskutWsElsewhere{} on the five sets it has not seen. SEFR CUT-ws1000 scores \WsSefr{} and
\SefrWsElsewhere. On so small a corpus, seen lines explain much, as two runs of LATTE-NC show. Trained on all 993 lines of Wisesight-1000, it scores
97.26 on WS160. Trained on 747 of them, without its 162 lines that equal or nearly repeat a WS160 line and an 84-line development split,
it scores only 88.04.

VISTEC gives a natural experiment. OSKut-scads scores \VistecSeen{} on the test lines inside its training split,
and \VistecUnseen{} on the \VistecCleanLines{} lines inside its own test split, a gap of \VistecGap. No other
system differs by more than \VistecOtherGap{} between the two groups. The other \VistecNeither{} lines are in
neither split, and every system scores lower on them. On those \VistecCleanLines{} lines,
\lw{} is above OSKut-scads by \VistecCleanDelta. About half of them hold a hashtag, which the files OSKut-scads
was trained on rarely do. On the \VistecNoHashLines{} lines without one, the difference is \VistecNoHashDelta.

\paragraph{Conventions decide.} A system scores far lower outside the convention it learnt. The four models
trained on BEST alone, namely DeepCut, SEFR CUT-best, AttaCut-SC and AttaCut-C, score \FourOnBest{} on our BEST
test lines and \FourElsewhere{} elsewhere. Seen lines explain little of this gap, as a held-out score shows
(Appendix~\ref{app:published}). On VISTEC, OSKut
scores \OskutWsVistec{} with its Wisesight engine, and \VistecUnseen{} with its VISTEC engine on the lines that
engine was not trained on. khammoon scores \KhammoonKhm{} on its own test and \KhammoonElsewhere{} elsewhere.
Even the three units of \lw{} differ by up to \UnitSpreadKhm{} points on one set
(Appendix~\ref{app:units}).

The dictionary tokenisers nlpO3 and \texttt{newmm} average \DictAvg, about \DictGap{} points below those four
models. Yet nlpO3 and TLTK score higher on ORCHID than any neural system of
Table~\ref{tab:main}, as the AttaCut documentation showed for \texttt{newmm}. ORCHID keeps long
units, as a dictionary does. The neural systems output \OrchidNeuralMore\%
more words than its gold, and the dictionary tools only \OrchidDictMore\%.

\section{Conclusion}

The idea of LATTE is to read a text at several granularities at once, and it survives on a CPU. \lw{} reads the
lattice as 13 bits per character, which add \NolexGainVistec{} points on VISTEC and \NolexGainAvgRound{} on
average. The larger gain comes from the teachers, whose labels on raw text carry each unit beyond its corpus. On
average, the student is as accurate as these labels, with 1.17M parameters against 635M for each teacher, on
one CPU thread.

Our revisit shows how much a ranking of Thai segmenters depends on conventions and on test lines seen in
training. We release the scripts, the scorer and the predictions, so that the next comparison of Thai
segmenters can start from them.

\section*{Limitations}

\paragraph{Coverage.} The seven sets cover encyclopedic, news, fiction, social-media, technical and instruction
text. Other domains, such as transcribed speech, are not covered. Our BEST and ORCHID splits are random at the line
level and are not official test sets. A model trained on the rest of such a corpus may score higher on them than on
new text.

\paragraph{Protocol.} Our rule that a space separates words changes some source tokens, so scores differ from those
on the original tokenisation. Upstream defects are kept as published (Appendix~\ref{app:sources}). Our intervals cover test sampling only.
They are not corrected for multiple comparisons, and they treat the lines of one document as independent.

\paragraph{Other tools.} They ran with their released models and default settings on one CPU thread.
Speeds were measured on one server, as the best round of each tool on one text. We state their licences as each project documented them in November 2026.

\paragraph{The model.} Its int8 activation scales are computed per batch. In another batch or alone, up to \BatchLinesMax\% of the
lines of a test set are segmented differently, and no score moves by more than \BatchFoneMax. Our scores come from
calls of 64 lines. Section numbers such as ``4.1'', other scripts, emoji sequences and some URLs can be split. The runtime is released with the model and in the \href{https://pypi.org/project/iapp-ai/}{\texttt{iapp-ai}} package on PyPI, but not the teachers or the training code. We did not train a teacher without the lattice, so its share in the labels is
not measured.

\paragraph{Selection.} Checkpoints and label mixtures were chosen on development data. The network size, the
training length and the dictionary features, however, come from an earlier phase. It compared
variants on test lines of BEST, Wisesight-1000 (which contains WS160), VISTEC and ORCHID.

The three units were designed after test scores of the teachers were known.
The rule that picks a unit for PUD, BEST and ORCHID was written after their \ushort{} test scores were known. At
\ushort, \lw{} is below DeepCut on PUD (\PudShortDeep). Table~\ref{tab:main} thus shows all units on all
sets, and our averages use \ushort.

\paragraph{Downstream use.} We do not test whether its output improves a language model. The studies we cite on
this concern Chinese and Japanese. For Thai, WangchanBERTa was pre-trained with word-level, syllable-level and
SentencePiece tokens to compare them~\citep{lowphansirikul2021wangchanberta}, and the khammoon paper compares
morphemes with words on three tasks~\citep{khammoon2026}.

\section*{Ethics and Licensing Statement}

\paragraph{Evaluation-only data.} We use BEST, ORCHID and the test set of khammoon for
evaluation, and the development splits of the first two to choose units. Its other non-commercial text
serves only to find repeats and for timing. None of these data trained the released model. LATTE-NC, the unreleased model of Section~\ref{sec:results}, also learnt from the training
splits of BEST and ORCHID, with a lattice from the non-commercial LEXiTRON dictionary.

\paragraph{Redistribution.} None of these is redistributed. The build script rebuilds them from their sources,
and their predictions hold word offsets only.

\paragraph{Personal data.} We removed dictionary entries that look like personal data (hashtags, handles,
e-mails, URLs, long numbers), but kept names.

\paragraph{Bias.} The model inherits the conventions of its teachers and the biases of its training text,
which is encyclopedia, social media and instructions.

\section*{Acknowledgements}

We thank SiamAI for computational resources, and the authors of khammoon, whose work rekindled the concept and
the motivation of this paper. The first author is deeply grateful to those who gave moral support during one of
the hardest times, when LATTE was being written. They believed in the work and in its author. An AI assistant was
used for coding, experiments, drafting, language editing and LaTeX. The authors reviewed and
verified all content and take full responsibility for it.
This research is financially supported under the Thammasat University's research fund, Center of Excellence in
Intelligent Informatics, Speech and Language Technology and Service Innovation (CILS), and Intelligent Informatics
and Service Innovation (IISI) Research Center, the Thailand Research Fund under grant number RTA6080013.

\bibliography{references}

\appendix
\onecolumn
\raggedbottom

\section{Systems Compared}\label{app:systems}

Table~\ref{tab:systems} lists the twelve systems of Table~\ref{tab:main}, which are nine tools, three of them with two models.
Each ran on the same lines and on one CPU thread, in one of five Python environments. It used its released model and default
settings, apart from the engines that Table~\ref{tab:systems} names.

The last column names the test sets whose lines are in the training corpus of the system. Their scores are in
parentheses in Table~\ref{tab:main}. For \lw{}, it names none, as it learnt from Wisesight-1000 without the
WS160 lines, and from CC~BY-SA rows of khammoon that hold no khm test document. The row of khammoon names BEST,
as its training text holds 147{,}999 sentences of the BEST training set, labelled as morphemes by a language model. Every line of our BEST test set is among them. DeepCut and AttaCut were each trained on 90\% of BEST, so they have seen most of our test lines.

SEFR CUT and OSKut refine the output of DeepCut with a second model trained on one corpus. The \texttt{scads}
engine of OSKut was trained on the split of VISTEC used in the OSKut paper, which contains \VistecShare{} of the
test lines of the corpus repository. Their TNHC engines and the LST20 engine of OSKut, which have no test set
here, were not run. Nor were their transfer-learned variants and the augmented one of OSKut. TLTK also ships a
list of BEST words for shorter units, which we did not run. As for licences, the code of DeepCut, AttaCut, SEFR CUT and OSKut is MIT-licensed. Their models were trained on
BEST (CC~BY-NC-SA~3.0), directly or through DeepCut. The code of khammoon is Apache-2.0, and its weights are
CC~BY-NC-SA~4.0, as stated on its model card.

\begin{table}[H]\centering\small\setlength{\tabcolsep}{4pt}
\begin{tabularx}{\linewidth}{@{}llll>{\raggedright\arraybackslash}Xl@{}}
\toprule
System & Version & Model & Parameters & Learnt from & Exposed sets \\
\midrule
LATTE-WordSeg-Lite & 0.1.2 & \texttt{long}, \texttt{short}, \texttt{minimal} & 1.17M & VISTEC, TUD, Wisesight-1000, khammoon & -- \\
khammoon & 0.1.0 & ONNX & 1.06M & text of BEST and WangchanThaiInstruct, its own LLM labels & BEST \\
DeepCut & 0.7.0.0 & -- & 0.54M & BEST & BEST \\
SEFR CUT & 1.1 & \texttt{best} & 0.54M + CRF & DeepCut + CRF trained on BEST & BEST \\
 &  & \texttt{ws1000} (default) & 0.54M + CRF & DeepCut + CRF trained on Wisesight-1000 & BEST, WS160 \\
AttaCut & 1.0.6 & \texttt{SC}, \texttt{C} & 0.16M, 0.17M & BEST & BEST \\
OSKut & 1.3 & \texttt{ws} (default) & 1.47M & DeepCut + network trained on Wisesight-1000 & BEST, WS160 \\
 &  & \texttt{scads} & 1.61M & DeepCut + network trained on VISTEC & BEST, VISTEC \\
TLTK & 1.11 & \texttt{colloc} (default) & -- & dictionary, collocation between syllables & -- \\
PyThaiNLP & 5.4.0 & \texttt{newmm} (default) & -- & dictionary & -- \\
PyThaiNLP nlpO3 & 1.4.0 & \texttt{newmm}, in Rust & -- & the same dictionary & -- \\
\bottomrule
\end{tabularx}

\caption{The systems compared, with their version, model or engine, parameters, training data and exposed test
sets. Parameters are counted from the released weights, and those of AttaCut are from its paper.}
\label{tab:systems}
\end{table}

Table~\ref{tab:speedfull} gives the exact speeds behind Table~\ref{tab:speed} and Figure~\ref{fig:tradeoff},
with peak memory. Every tool segments the same 768 lines on one thread of an Intel Xeon Platinum 8468, with model
loading left out. The lines are the validation part of the development set of khammoon (150{,}894 characters),
used here for timing only. A run takes the best of up to three rounds, and the best run of a tool counts. Over its
two runs, \lw{} ran at \SpeedLite{} characters per second, and khammoon at \SpeedKhammoon{} over its four. Every thread pool was set to one thread, but the TensorFlow tools still used
\TfCores{} cores (CPU time over wall time). DeepCut, SEFR CUT and OSKut ran on TensorFlow 2.15, AttaCut on
PyTorch 2.5, and khammoon and \lw{} on ONNX Runtime 1.32 and 1.31.

khammoon and SEFR CUT take all lines in one call, and the other tools one line per call. DeepCut makes one
prediction call per line, which bounds its speed. The row per line of \lw{} is the lower of two runs of earlier
models of its architecture, each right after its batched rounds. In a later session, the released model ran at
\SpeedLiteBatchR{} characters per second batched and at \SpeedLiteLineR{} per line, in rounds that alternate.
DeepCut ran at \SpeedDeepcutR{} in that session, \SpeedLineVsDeepcutR{} times slower, and khammoon at
\SpeedKhammoonLine{} per line. \lw{} runs as an int8 graph and khammoon in float32. In another session, our
network ran at \SpeedLiteFp{} characters per second in float32, its int8 graph at \SpeedLiteLater{} and khammoon at
\SpeedKhammoonLater.

The second engines of SEFR CUT and OSKut were not timed. They re-score every position, where the engines timed
re-score 5\% and 33\% of them, so they would be slower. The teacher was timed for one convention. On the CPU thread, it
read the first 60 lines of this text. On the GPU, it read the text 20 times over, as when it labelled the raw text.
Its layers were compiled, and six CPU processes prepared its input. Its \ushort{} labels mix two conventions, so they take two passes and
twice the time.

\begin{table}[H]\centering\small
\begin{minipage}{0.72\linewidth}
\begin{tabular*}{\linewidth}{@{\extracolsep{\fill}}lrr@{}}
\toprule
System & Characters per second & Peak memory (MB) \\
\midrule
PyThaiNLP nlpO3 & 5{,}056{,}595 & 90 \\
PyThaiNLP newmm & 936{,}850 & 101 \\
\textbf{LATTE-WordSeg-Lite}, batched & 104{,}709 & 136 \\
AttaCut-C & 99{,}619 & 437 \\
AttaCut-SC & 78{,}500 & 433 \\
\textbf{LATTE-WordSeg-Lite}, per line & 59{,}588 & -- \\
khammoon & 43{,}075 & 268 \\
DeepCut & 2{,}576 & 648 \\
SEFR CUT-best & 1{,}887 & 705 \\
OSKut-ws & 1{,}082 & 700 \\
TLTK & 345 & 413 \\
\addlinespace LATTE teacher, one CPU thread & 66 & 7{,}271 \\
LATTE teacher, one H100 GPU & 92{,}076 & -- \\
\bottomrule
\end{tabular*}

\end{minipage}
\caption{Speed on one CPU thread and peak memory, as measured.}
\label{tab:speedfull}
\end{table}

PyThaiNLP wraps most of these tools. nlpO3 is an adaptation of its \texttt{newmm} in Rust, run here with the
same dictionary. Their outputs differ by up to \NlpoNewmmMax{} points on a set. PyThaiNLP also has
four more word tokenisers of its own (Table~\ref{tab:pythainlp}), all run with its default dictionary.
\texttt{newmm-safe} guards \texttt{newmm} against inputs that take it very long. \texttt{longest} and \texttt{mm} use longest
and maximum matching. \texttt{nercut} merges the output of a named-entity tagger. None comes closer to the neural
systems than \texttt{newmm}. Its ICU tokeniser was not run, as PyICU, which it needs, was not available
in the Python 3.11 environment of these tools.

\begin{table}[H]\centering\small
\begin{tabular*}{\linewidth}{@{\extracolsep{\fill}}lrrrrrrrr@{}}
\toprule
Engine & TUD & PUD & WS160 & VISTEC & BEST & ORCHID & khm & Avg \\
\midrule
\texttt{newmm} (Table~\ref{tab:main}) & 69.64 & 67.96 & 73.35 & 76.88 & 69.13 & 79.17 & 60.18 & 70.90 \\
\texttt{newmm-safe} & 69.66 & 67.96 & 73.35 & 76.87 & 69.13 & 79.17 & 60.18 & 70.90 \\
\texttt{longest} & 68.93 & 67.21 & 72.27 & 75.64 & 68.95 & 79.16 & 61.04 & 70.46 \\
\texttt{mm} & 62.00 & 60.62 & 65.23 & 66.98 & 61.82 & 73.05 & 54.23 & 63.42 \\
\texttt{nercut} & 62.00 & 61.67 & 64.39 & 65.95 & 62.53 & 72.93 & 54.41 & 63.41 \\
\bottomrule
\end{tabular*}

\caption{The word tokenisers of PyThaiNLP 5.4.0 (word F1, \%), scored as in Table~\ref{tab:main}.}
\label{tab:pythainlp}
\end{table}

\section{Model and Training}\label{app:training}

\paragraph{Our LATTE and the released code.} A lattice word is embedded from the encoder states of its first and
last characters and their mean, with a hashed identity. The released LATTE code runs the encoder once per word
instead. The bidirectional lattice attention works on batched neighbour lists, not on one graph per sentence. On CTB6
with bert-base-chinese, a training step on 16 sentences takes 0.071~s in fp32 (0.062~s in bf16), against 0.257~s
for the released code in fp32 on the same GPU. A word that only the training data contains is hidden from the
lattice of the sentences it comes from.

We also measured what the lattice and its graph layers add. On our BEST split before de-duplication, our LATTE with multilingual BERT scores 97.69 word F1. The same encoder
with a CRF and no lattice scores 96.81. Trained on longer chunks, our LATTE scores 97.65 with its two graph
layers and 97.45 without them. Each number is one run. These models are for research only, as BEST is licensed for non-commercial use, and none of
them is a teacher of \lw{}.

\paragraph{Data and settings.} Table~\ref{tab:data} lists every ingredient of the release with its licence.
Table~\ref{tab:settings} gives the settings of the teachers and of the student. Development splits only select
checkpoints and mixture weights. They hold 4{,}000 VISTEC lines, 362 TUD sentences, 84 Wisesight lines and 251
khammoon documents. Teacher~1 was selected on a sample of 3{,}000 of the VISTEC lines. The student learns fast. Its mean development F1 is 90.89 after 2{,}000 steps, about one pass over the
labelled text. It is 92.16 after 10{,}000 steps and 92.79 at its best step, 54{,}000.

\paragraph{Dictionary and decoding.} The released dictionary has 177{,}385 words. It is the union of the word
lists and the training vocabulary of Teacher~1 (190{,}037 words), minus 12{,}652 entries that look like personal
data. Removing them changes no test score by more than 0.11. The six bands of word length of the lattice bits are
2, 3, 4, 5--6, 7--9 and 10--16 characters. The 13 bits of a character select one of 8{,}192 embeddings. In decoding, a word may not start at a combining mark or a following vowel, nor right after a
leading vowel. Runs of ASCII letters or of digits are never split. Every \ulong{} boundary is a \ushort{}
boundary, as the \ushort{} logit is the maximum of both. \umin{} is decoded on its own, because khammoon keeps
together some punctuation that \ushort{} splits. On the khammoon development documents, \umin{} reaches a boundary
F1 of 99.80 on its own, against 99.24 when they are nested in those of \ushort.

\begin{table}[H]\centering\small
\begin{tabularx}{\linewidth}{@{}>{\raggedright\arraybackslash}Xlll@{}}
\toprule
Data & Used for & Size & Licence \\
\midrule
VISTEC, training split & Teacher~1, \texttt{long} & 36,000 lines & CC BY-SA 3.0 \\
UD Thai-TUD, training split & Teacher~1, \texttt{short} & 2,902 sentences & CC BY-SA 4.0 \\
Wisesight-1000 without the WS160 lines & Teacher~1, \texttt{short} & 747 lines & CC0 1.0 \\
khammoon data, rows under CC~BY-SA & Teacher~2, \texttt{minimal} & 4,763 documents & CC BY-SA 4.0 \\
\addlinespace Thai Wikipedia, dump of 2023-11-01 & soft labels & 60.0M characters & CC BY-SA \\
Wisesight sentiment, train and validation & soft labels & 1.9M characters & CC0 1.0 \\
Training documents of Teacher~2 & soft labels & 7.2M characters & CC BY-SA 4.0 \\
\addlinespace Word lists of PyThaiNLP, ICU and Volubilis 23.1 & dictionary & 130,318 words & CC0, Unicode, CC BY-SA 4.0 \\
Training vocabulary of Teacher~1 & dictionary & 100,860 words & as its corpora \\
\addlinespace XLM-RoBERTa-large & teacher encoder & -- & MIT \\
\bottomrule
\end{tabularx}

\caption{Everything \lw{} was trained on or built from, with the size used and the licence.}
\label{tab:data}
\end{table}

\begin{table}[H]\centering\small
\begin{tabular*}{\linewidth}{@{\extracolsep{\fill}}lll@{}}
\toprule
 & LATTE teachers & \lw{} student \\
\midrule
Encoder & XLM-RoBERTa-large & 64-d characters; convolutions 3, 5, 7 ($\times$128) \\
Layers on top & two lattice-attention layers, 8 heads & two BiLSTM layers, 128 units per direction \\
Lattice & words of up to 16 characters, as nodes & 13 bits per character, as 32 dimensions \\
Output & one CRF per training corpus & one logit per unit \\
Parameters & 634.6M & 1.17M (1,168,771) \\
\addlinespace Labels & corpus labels (Table~\ref{tab:data}) & teacher probabilities, 69.1M characters \\
Loss & CRF likelihood & binary cross-entropy, all three units \\
Optimiser & AdamW, weight decay 0.01 & AdamW, weight decay 0.01 \\
Learning rate & $1.5\times10^{-5}$ (encoder, output layer), $2\times10^{-4}$ (other) & $2\times10^{-3}$ \\
Schedule & linear decay; warm-up 6\% and 5\% & cosine decay; warm-up 500 steps \\
Length & 6 and 5 epochs & 60K steps \\
Batch & up to 4,096 tokens & up to 32,768 characters \\
Regularisation & dropout 0.1; gradient clipping 1.0 & dropout 0.2; gradient clipping 1.0 \\
Selection & development F1, every epoch & mean development F1, every 2,000 steps \\
\addlinespace Time, one H100 & 47 and 26 minutes & 51 minutes labelling, 57 training \\
Released as & not released & one int8 ONNX graph, 3.0~MB \\
\bottomrule
\end{tabular*}

\caption{Settings of the LATTE teachers and of the \lw{} student, whose best checkpoint is step 54{,}000. Where
two values are given, the first is for Teacher~1 and the second for Teacher~2.}
\label{tab:settings}
\end{table}

\section{Test Sets}\label{app:sources}

Table~\ref{tab:sources} pins the source of each test set, and the benchmark lists the SHA-256 of every source file
in \texttt{sources.json}. Our splits of BEST and ORCHID are drawn with a fixed seed. As for the repeats of
Section~\ref{sec:bench}, a test line is one when, ignoring whitespace, it equals a training line. It is one too when at least 80\% of
its 12-character shingles (every fourth) occur in the training text. BEST
and ORCHID are checked against their train and dev splits, and khm against the training text of khammoon. That
text is three more files of its repository, at the same pinned revision.

\begin{table}[H]\centering\small
\begin{tabular*}{\linewidth}{@{\extracolsep{\fill}}lllll@{}}
\toprule
Set & Host & Repository & Revision & Licence \\
\midrule
TUD & GitHub & \texttt{UniversalDependencies/UD\_Thai-TUD} & \texttt{d4cb5fd3} & CC BY-SA 4.0 \\
PUD & GitHub & \texttt{UniversalDependencies/UD\_Thai-PUD} & \texttt{4e0e10fd} & CC BY-SA 3.0 \\
WS160 & GitHub & \texttt{PyThaiNLP/wisesight-sentiment} & \texttt{6995f993} & CC0 1.0 \\
VISTEC & GitHub & \texttt{mrpeerat/OSKut} & \texttt{be9e16ac} & CC BY-SA 3.0 \\
BEST & Hugging Face & \texttt{nectec/best2009} & \texttt{685fffc4} & CC BY-NC-SA 3.0 \\
ORCHID & GitHub & \texttt{wannaphong/corpus\_mirror} & \texttt{orchid-v1.0 (tag)} & CC BY-NC-SA 3.0 \\
khm & Hugging Face & \texttt{typhoon-ai/khammoon-data} & \texttt{400ded89} & CC BY-NC 4.0 \\
\bottomrule
\end{tabular*}

\caption{Source, pinned revision and licence of each test set.}
\label{tab:sources}
\end{table}

\begin{table}[H]
\begin{minipage}[t]{0.585\linewidth}\vspace{0pt}\setlength{\parindent}{1em}
\noindent Table~\ref{tab:exposure} counts the test lines with a copy or near-copy in the text that \lw{} learnt
from. Such a line equals a training line, lies inside one, or shares at least 80\% of its shingles with one. These
lines are at most \ExposedMaxOtherStudent{} of any other set, against \ExposedKhmStudent{} of khm, mostly short headings
and templates of the khammoon documents. With the training data of the teachers added, khm has \ExposedKhm{} of such lines.

Removing the lines of the Any column changes no paired difference of Appendix~\ref{app:units} by more than
\ExposedChange{} points, and reverses the sign of none. Between \lw{} and any other system on any set, the largest
change is \ExposedChangeAll{} points, on WS160 against TLTK, a set of only 160 lines.
\end{minipage}\hfill
\begin{minipage}[t]{0.365\linewidth}\vspace{0pt}\small
{\centering\begin{tabular*}{\linewidth}{@{\extracolsep{\fill}}lrrr@{}}
\toprule
Set & Lines & Student & Any \\
\midrule
TUD & 363 & 3 & 5 \\
PUD & 1{,}000 & 0 & 0 \\
WS160 & 160 & 5 & 5 \\
VISTEC & 10{,}000 & 1 & 157 \\
BEST & 13{,}072 & 13 & 15 \\
ORCHID & 1{,}747 & 9 & 11 \\
khm & 13{,}731 & 1{,}099 & 1{,}246 \\
khmdd & 10{,}522 & 28 & 29 \\
\bottomrule
\end{tabular*}
\par}
\caption{Test lines with a copy or near-copy in the training text of the student (Student), or in that text and
the training data of the two teachers (Any).}
\label{tab:exposure}
\end{minipage}
\end{table}

Table~\ref{tab:defects} lists what we kept as published, and none of it changes a ranking. We also checked our
own code twice. We rebuilt every test file with independent code, and we recomputed the \ScorerScores{} scores of
Table~\ref{tab:main} with a second scorer written from scratch. The two scorers agree within \ScorerAgree{} points
for every system but TLTK. TLTK rewrites some characters of its input. Our scorer still credits it with the words
that it cut, the second one does not, and the two then differ by at most \ScorerAgreeTltk{} points.

\begin{table}[H]\centering\small
\begin{tabularx}{\linewidth}{@{}>{\raggedright\arraybackslash}Xl@{}}
\toprule
Kept as published & Where \\
\midrule
The text of the treebank differs from the original sentence (in the third by eleven added spaces) & PUD (ids 72, 611 and 979) \\
A space between every word & TUD (id 300) \\
Literal \texttt{<POEM>} markup; a byte-order mark & BEST (54 lines; 4 lines) \\
Word lines drop what follows a slash (\texttt{TCP/IP} becomes \texttt{TCP}); a literal \texttt{\%@PUNC} & ORCHID (all such words; 1 line) \\
A gold word that is only a zero-width space & khm (16 words), WS160 (1) \\
A gold word that starts with a combining mark, which no segmenter that keeps grapheme clusters can produce & VISTEC (19 words), ORCHID (29), BEST (5) \\
\addlinespace Our rule, not a defect, splits a source token with a space inside & BEST (2,214 tokens), VISTEC (1,121), WS160 (19) \\
\bottomrule
\end{tabularx}

\caption{Defects of the sources, kept as published, and the effect of our rule that a space separates words. An
id is the \texttt{id} field of a line in our test file of that set, counted from zero.}
\label{tab:defects}
\end{table}

\section{Published Scores, Reproduced}\label{app:published}

Thai segmentation papers score in three ways, which give different numbers for the same output. SEFR CUT and
OSKut count over the whole test file and keep the spaces. Every character is a word start or not, and every token,
a space included, is a word. AttaCut and the PyThaiNLP benchmark remove the spaces and average over lines.
khammoon scores the cuts between two non-space characters, as we do.

Table~\ref{tab:published} applies each way to our stored outputs, on the test file of each paper. For VISTEC,
this is the split that the OSKut paper used, which the README of its repository links. For Wisesight, it is
Wisesight-1000 and WS160 as published. OSKut and DeepCut are reproduced to the second decimal of the OSKut paper,
but for the word-level score of OSKut on WS160, which is 0.01 lower. SEFR CUT is reproduced in the row of its own
paper. Four of the five tools of the khammoon table match it to the decimal it
prints, and the fifth, AttaCut-SC, is 0.4 below it. Only TLTK, and the \texttt{best} engine of SEFR CUT, have no
published score to check against.
AttaCut matches the benchmark of its authors at the two digits printed there, and differs from the other papers by
up to 1.8. The SEFR CUT paper trained its own AttaCut-SC model. \texttt{newmm} differs by 2.1 on the PyThaiNLP
benchmark. These numbers date from 2019 to 2021, and we do not know their releases. Published numbers
disagree too, as the SEFR CUT paper lists DeepCut 0.3 higher at character level than the OSKut paper and our run.

One published score is on lines that no tool has seen. On the official test set of BEST, the AttaCut paper reports
a word F1 of \BestHeldDeepcut{} for DeepCut, as a mean over lines~\citep{chormai2019attacut}. Scored that way, our
BEST test lines give \BestOursDeepcut. \texttt{newmm}, which learns nothing from BEST, moves as much, from
\BestHeldNewmm{} to \BestOursNewmm. So the lines that DeepCut has seen explain little of the gap between its BEST
score and its other scores, which is mostly a matter of convention.

\begin{table}[H]\centering\small
\begin{tabular*}{\linewidth}{@{\extracolsep{\fill}}lllrr@{}}
\toprule
Test file and scoring & System & Source & Published & Ours \\
\midrule
WS160, spaces kept & DeepCut & OSKut paper & 93.47 / 84.03 & 93.47 / 84.03 \\
 & DeepCut & SEFR CUT paper & 93.8 / 84.0 & 93.47 / 84.03 \\
 & AttaCut-SC & OSKut paper & 93.50 / 84.04 & 93.37 / 83.58 \\
 & AttaCut-SC & SEFR CUT paper & 93.5 / 84.0 & 93.37 / 83.58 \\
 & SEFR CUT-ws1000 & SEFR CUT paper & 95.2 / 87.4 & 95.17 / 87.40 \\
 & OSKut-ws & OSKut paper & 96.67 / 91.51 & 96.67 / 91.50 \\
\addlinespace VISTEC, the split of OSKut, spaces kept & OSKut-scads & OSKut paper & 97.36 / 92.91 & 97.36 / 92.91 \\
 & DeepCut & OSKut paper & 92.77 / 81.78 & 92.77 / 81.78 \\
 & AttaCut-SC & OSKut paper & 91.47 / 79.31 & 92.32 / 81.12 \\
\addlinespace Wisesight-1000, mean over lines & PyThaiNLP newmm & AttaCut benchmark & 90 / 74 & 90 / 75 \\
 & DeepCut & AttaCut benchmark & 92 / 81 & 92 / 81 \\
 & AttaCut-C & AttaCut benchmark & 92 / 80 & 92 / 80 \\
 & AttaCut-SC & AttaCut benchmark & 93 / 81 & 93 / 81 \\
\addlinespace WS160, mean over lines & PyThaiNLP newmm & PyThaiNLP benchmark & 72.48 & 74.53 \\
 & DeepCut & PyThaiNLP benchmark & 80.42 & 80.42 \\
\addlinespace khm, boundary F1 & PyThaiNLP newmm & khammoon paper & 80.2 & 80.15 \\
 & PyThaiNLP nlpO3 & khammoon paper & 80.5 & 80.48 \\
 & DeepCut & khammoon paper & 92.1 & 92.07 \\
 & AttaCut-SC & khammoon paper & 92.5 & 92.11 \\
 & khammoon & khammoon paper & 99.2 & 99.22 \\
\bottomrule
\end{tabular*}

\caption{Published scores and our runs, scored as in each source, in \%. Where two numbers are given, they are
the character-level and the word-level F1. The sources are Table~3 of the SEFR CUT~\citep{limkonchotiwat2020sefr} and OSKut~\citep{limkonchotiwat2021oskut}
papers, the benchmark page of the AttaCut documentation, whose word-level values are Table~2 of its
paper~\citep{chormai2019attacut}, the file \texttt{tokenization-benchmark.md} of the PyThaiNLP
repository~\citep{phatthiyaphaibun2023pythainlp} and Table~6 of the khammoon paper~\citep{khammoon2026}. The
khammoon paper scores whole documents, and we score the lines of these documents.}
\label{tab:published}
\end{table}

\section{Results in Full}\label{app:units}

Table~\ref{tab:studentsets} gives the rows of Table~\ref{tab:student} on the test sets of the corpora each unit
was trained on, with the second run of the student. Over the three units and the seven sets,
the two runs differ by at most \SeedMaxExact, for \ushort{} on ORCHID. Their means over the seven sets differ by at most \SeedMeanMax. On the 160 lines of WS160, the
corpus-trained network differs from the student of the same seed by \GoldWsCi. Table~\ref{tab:fidelity} compares the cuts of the student
with those of its teachers. Where they differ, gold sides with the teachers on VISTEC (\FidVistec) and with the
student on PUD, BEST and khm (\FidStudentRange). Table~\ref{tab:significance} gives the paired differences behind Section~\ref{sec:results}. \lw{} is at the unit
of each set, against the systems named in the text and against DeepCut on every set.

\begin{table}[H]\centering\small
\begin{minipage}{0.66\linewidth}
\begin{tabular*}{\linewidth}{@{\extracolsep{\fill}}lrrrr@{}}
\toprule
Test set & VISTEC & TUD & WS160 & khm \\
Unit & \texttt{long} & \texttt{short} & \texttt{short} & \texttt{minimal} \\
\midrule
LATTE teachers & 93.66 & 90.20 & 88.09 & 98.36 \\
\textbf{\lw{}} & 92.77 & 90.27 & 88.32 & 98.44 \\
\lw{}, second run & 92.59 & 90.10 & 88.29 & 98.43 \\
\addlinespace Single-unit students & 93.04 & 90.30 & 88.67 & 98.41 \\
Corpus-trained networks & 93.62 & 91.50 & 86.15 & 98.29 \\
\addlinespace Student without lattice bits & 91.51 & 90.16 & 88.72 & 98.45 \\
\bottomrule
\end{tabular*}

\end{minipage}
\caption{Table~\ref{tab:student} on the test sets of the corpora of each unit, with the second run (word F1, \%).}
\label{tab:studentsets}
\end{table}

\begin{table}[H]\centering\small
\begin{tabular*}{\linewidth}{@{\extracolsep{\fill}}lrrrrrr@{}}
\toprule
 & & Identical & Boundary F1 & Differing & \multicolumn{2}{c}{Gold sides with} \\
Set & Unit & lines (\%) & between them & cuts & teachers (\%) & student (\%) \\
\midrule
TUD & \texttt{short} & 59.5 & 98.14 & 237 & 50.2 & 49.8 \\
PUD & \texttt{long} & 49.8 & 97.99 & 738 & 35.0 & 65.0 \\
WS160 & \texttt{short} & 65.6 & 97.55 & 124 & 47.6 & 52.4 \\
VISTEC & \texttt{long} & 28.9 & 98.55 & 15{,}968 & 64.5 & 35.5 \\
BEST & \texttt{short} & 42.6 & 98.18 & 15{,}253 & 43.5 & 56.5 \\
ORCHID & \texttt{long} & 62.7 & 97.37 & 997 & 49.9 & 50.1 \\
khm & \texttt{minimal} & 92.8 & 99.80 & 1{,}353 & 39.6 & 60.4 \\
\addlinespace All seven & & 57.7 & 98.72 & 34{,}670 & 53.1 & 46.9 \\
\bottomrule
\end{tabular*}

\caption{The student against the labels of its teachers, each set at its unit. The table gives the lines that
both segment identically, the boundary F1 between their cuts and, where their cuts differ, how often gold agrees
with each.}
\label{tab:fidelity}
\end{table}

\begin{table}[H]\centering\small
\begin{tabular*}{\linewidth}{@{\extracolsep{\fill}}llrlrr@{}}
\toprule
Set & Unit & Ours & Versus & F1 & $\Delta$ [95\% interval] \\
\midrule
TUD & \texttt{short} & 90.27 & khammoon & 85.52 & $+4.75$ [$+3.19$, $+6.21$] \\
PUD & \texttt{long} & 86.18 & DeepCut & 85.62 & $+0.56$ [$-0.15$, $+1.25$] \\
PUD & \texttt{long} & 86.18 & SEFR CUT-best & 85.55 & $+0.63$ [$-0.06$, $+1.32$] \\
WS160 & \texttt{short} & 88.32 & OSKut-scads & 83.20 & $+5.12$ [$+2.88$, $+7.33$] \\
WS160 & \texttt{short} & 88.32 & SEFR CUT-ws1000 & (86.69) & $+1.64$ [$-0.50$, $+3.98$] \\
WS160 & \texttt{short} & 88.32 & OSKut-ws & (91.28) & $-2.96$ [$-4.99$, $-1.05$] \\
VISTEC & \texttt{long} & 92.77 & OSKut-scads & (89.69) & $+3.08$ [$+2.89$, $+3.27$] \\
VISTEC & \texttt{long} & 92.77 & SEFR CUT-best & 80.39 & $+12.38$ [$+12.21$, $+12.56$] \\
BEST & \texttt{short} & 86.97 & SEFR CUT-best & (97.55) & $-10.58$ [$-10.74$, $-10.43$] \\
ORCHID & \texttt{long} & 79.77 & PyThaiNLP nlpO3 & 80.63 & $-0.86$ [$-1.63$, $-0.11$] \\
ORCHID & \texttt{long} & 79.77 & TLTK & 80.68 & $-0.91$ [$-1.72$, $+0.00$] \\
khm & \texttt{minimal} & 98.44 & khammoon & 98.33 & $+0.12$ [$+0.08$, $+0.16$] \\
VISTEC, 1{,}821 clean lines & \texttt{long} & 93.19 & OSKut-scads & 90.37 & $+2.82$ [$+2.59$, $+3.09$] \\
VISTEC, 874 of them without hashtag & \texttt{long} & 93.40 & OSKut-scads & 92.45 & $+0.95$ [$+0.72$, $+1.19$] \\
\addlinespace
TUD & \texttt{short} & 90.27 & DeepCut & 85.50 & $+4.76$ [$+3.74$, $+5.85$] \\
WS160 & \texttt{short} & 88.32 & DeepCut & 83.13 & $+5.19$ [$+3.31$, $+7.41$] \\
VISTEC & \texttt{long} & 92.77 & DeepCut & 80.12 & $+12.65$ [$+12.48$, $+12.82$] \\
BEST & \texttt{short} & 86.97 & DeepCut & (97.54) & $-10.57$ [$-10.73$, $-10.43$] \\
ORCHID & \texttt{long} & 79.77 & DeepCut & 76.68 & $+3.09$ [$+2.45$, $+3.71$] \\
khm & \texttt{minimal} & 98.44 & DeepCut & 82.81 & $+15.63$ [$+15.47$, $+15.80$] \\
\addlinespace
PUD & \texttt{short} & 84.66 & DeepCut & 85.62 & $-0.96$ [$-1.54$, $-0.36$] \\
khm & \texttt{short} & 82.29 & DeepCut & 82.81 & $-0.52$ [$-0.67$, $-0.37$] \\
\addlinespace
Average, all seven sets & \texttt{short} & 85.65 & SEFR CUT-best & 84.52 & $+1.13$ [$+0.80$, $+1.52$] \\
Average, commercial-use sets & \texttt{short} & 87.80 & SEFR CUT-ws1000 & 83.78 & $+4.01$ [$+3.43$, $+4.69$] \\
Average, PUD and ORCHID & \texttt{short} & 81.88 & SEFR CUT-best & 81.16 & $+0.72$ [$+0.36$, $+1.10$] \\
\addlinespace khm, the protocol of khammoon & \texttt{minimal} & 99.28 & khammoon & 99.23 & $+0.056$ [$+0.032$, $+0.079$] \\
\bottomrule
\end{tabular*}

\caption{\lw{} minus other systems in F1 points, with 95\% intervals from 1{,}000 paired bootstrap
resamples of lines. Parentheses are as in Table~\ref{tab:main}. The clean lines are the VISTEC test lines
inside the test split of OSKut-scads. The last row is boundary F1 on whole documents, the protocol of
khammoon (10{,}000 resamples).}
\label{tab:significance}
\end{table}

Table~\ref{tab:boundary} is Table~\ref{tab:main} with boundary
F1, the second metric. With \ushort{} on every set, the average of \lw{} is \BAvgWord{}, against \BAvgSefr{}
for SEFR CUT-best (\BAvgDelta), the second highest. Counting the cuts next to a space too would add \CharAllLo{} to \CharAllHi{}
points to boundary F1 and reorder three pairs of systems, each within 0.1.

\begin{table}[H]\centering\small\setlength{\tabcolsep}{3.5pt}
\begin{tabular*}{\linewidth}{@{\extracolsep{\fill}}lrrrrrrrrr@{}}
\toprule
 & \multicolumn{4}{c}{Commercial-use licence} & \multicolumn{3}{c}{Non-commercial licence} & \multicolumn{2}{c}{Average} \\
\cmidrule(lr){2-5}\cmidrule(lr){6-8}\cmidrule(l){9-10}
System & TUD & PUD & WS160 & VISTEC & BEST & ORCHID & khm & All & Commercial \\
\midrule
\textbf{LATTE-WordSeg-Lite} & & & & & & & & & \\
\quad \texttt{long} & 92.92 & \textbf{94.23} & 92.13 & \textbf{96.61} & 92.18 & \textbf{89.42} & 86.23 & 91.96 & \underline{93.97} \\
\quad \texttt{short} & \textbf{95.90} & 93.61 & \underline{93.98} & 94.46 & 94.27 & 88.92 & 91.98 & \textbf{93.30} & \textbf{94.49} \\
\quad \texttt{minimal} & 93.89 & 91.69 & 92.11 & 88.51 & 91.85 & 83.19 & \textbf{99.28} & 91.50 & 91.55 \\
\midrule
khammoon & 93.93 & 91.46 & 91.38 & 88.07 & (91.91) & 83.24 & \underline{99.22} & 91.32 & 91.21 \\
DeepCut & \underline{93.97} & \underline{94.01} & 91.39 & 90.25 & \underline{(98.91)} & 87.47 & 92.07 & 92.58 & 92.41 \\
SEFR CUT-best & 93.93 & 93.98 & 91.38 & 90.39 & \textbf{(98.92)} & 87.54 & 92.08 & \underline{92.60} & 92.42 \\
SEFR CUT-ws1000 & 93.80 & 93.71 & (93.62) & 89.44 & (97.44) & 86.83 & 92.74 & 92.51 & 92.64 \\
AttaCut-SC & 93.80 & 93.87 & 91.37 & 89.62 & (98.30) & 87.43 & 92.11 & 92.36 & 92.16 \\
AttaCut-C & 93.68 & 93.49 & 90.66 & 88.79 & (98.04) & 87.03 & 91.87 & 91.94 & 91.65 \\
OSKut-ws & 88.90 & 85.90 & \textbf{(95.78)} & 85.94 & (88.55) & 80.42 & 90.94 & 88.06 & 89.13 \\
OSKut-scads & 90.13 & 92.79 & 90.84 & \underline{(94.57)} & (90.35) & 85.76 & 85.05 & 89.93 & 92.08 \\
TLTK & 89.37 & 88.06 & 87.44 & 89.42 & 88.71 & \underline{89.24} & 83.01 & 87.89 & 88.57 \\
PyThaiNLP nlpO3 & 86.25 & 85.23 & 85.15 & 87.32 & 85.95 & 88.97 & 80.48 & 85.62 & 85.99 \\
PyThaiNLP newmm & 86.21 & 85.27 & 85.84 & 88.40 & 85.79 & 88.14 & 80.15 & 85.69 & 86.43 \\
\bottomrule
\end{tabular*}

\caption{Boundary F1 (\%). This is Table~\ref{tab:main} with the second metric, the cuts between two characters,
without those next to a space. Columns, bold, underline and parentheses are as in Table~\ref{tab:main}.}
\label{tab:boundary}
\end{table}

Table~\ref{tab:units} gives \lw{} at every unit on every test set, and the held-out scores that chose the unit
for PUD, BEST and ORCHID. Table~\ref{tab:khm} gives the metric of khammoon on its test set.

\begin{table}[H]\small
\begin{minipage}[t]{0.455\linewidth}\vspace{0pt}
{\centering\begin{tabular*}{\linewidth}{@{\extracolsep{\fill}}lrrr@{}}
\toprule
Test set & \texttt{long} & \texttt{short} & \texttt{minimal} \\
\midrule
TUD & 83.71 & \textbf{90.27} & 85.36 \\
PUD & \textbf{86.18} & 84.66 & 80.34 \\
WS160 & 85.51 & \textbf{88.32} & 84.08 \\
VISTEC & \textbf{92.77} & 87.94 & 76.48 \\
BEST & 82.09 & \textbf{86.97} & 82.09 \\
ORCHID & \textbf{79.77} & 79.10 & 67.86 \\
khm & 71.54 & 82.29 & \textbf{98.44} \\
Average, all seven & 83.08 & 85.65 & 82.09 \\
\midrule
Held-out data & \texttt{long} & \texttt{short} & \texttt{minimal} \\
\midrule
BEST, dev split & 82.09 & \textbf{86.81} & 81.85 \\
ORCHID, dev split & \textbf{79.34} & 78.53 & 66.65 \\
PUD, half A & \textbf{85.76} & 84.20 & 80.13 \\
PUD, half B & \textbf{86.60} & 85.12 & 80.55 \\
\bottomrule
\end{tabular*}
\par}
\caption{The three units (word F1, \%). Bold marks the unit taken for a set in comparisons, and the unit chosen
on held-out data. PUD has no other split, so each half of its test set chooses for the other half.}
\label{tab:units}
\end{minipage}\hfill
\begin{minipage}[t]{0.50\linewidth}\vspace{0pt}
{\centering\begin{tabular*}{\linewidth}{@{\extracolsep{\fill}}lrrr@{}}
\toprule
System & khm & khmdd & Documents \\
\midrule
\textbf{LATTE-WordSeg-Lite} & 99.28 & 99.27 & 99.28 \\
khammoon & 99.22 & 99.20 & 99.23 \\
DeepCut & 92.07 & 92.23 & -- \\
SEFR CUT-best & 92.08 & 92.24 & -- \\
SEFR CUT-ws1000 & 92.74 & 92.89 & -- \\
AttaCut-SC & 92.11 & 92.26 & -- \\
AttaCut-C & 91.87 & 92.00 & -- \\
OSKut-ws & 90.94 & 91.03 & -- \\
OSKut-scads & 85.05 & 85.41 & -- \\
TLTK & 83.01 & 83.32 & -- \\
PyThaiNLP nlpO3 & 80.48 & 80.87 & -- \\
PyThaiNLP newmm & 80.15 & 80.53 & -- \\
\bottomrule
\end{tabular*}
\par}
\caption{The metric of khammoon (boundary F1, \%) on its test set. The khm column scores lines, and khmdd the
lines that do not repeat its training text. The Documents column uses the evaluation code of khammoon. \lw{} is
at its \umin{} unit.}
\label{tab:khm}
\end{minipage}
\end{table}

\paragraph{Several readings.} \lw{} can also list the $k$ most probable segmentations of a line. They are the
$k$ best paths through a lattice of the dictionary words in the line and the words of its three units. The start
probabilities of the student score each path. For \emph{taklom} (Figure~\ref{fig:graph}), it lists
\emph{ta\textbar klom} (`round eyes') first and \emph{tak\textbar lom} (`out in the wind') third. The lists are made one line at a time, where the
average word F1 is \NbestOne{}, with \ushort{} on every set. Taking for each line the best of its top 3 raises it to
\NbestThree. The top 5 and 10 give \NbestFive{}
and \NbestTen. These are upper bounds, since the gold standard picks the reading. On TUD, the first reading is
right for the whole line in \NbestTudOne\% of the lines, and one of the top 3 in \NbestTudThree\%.

\end{document}